\documentclass[sigconf]{acmart}

\usepackage{threeparttable}
\usepackage{csquotes}
\usepackage{multicol}
\usepackage{booktabs}
\usepackage{graphicx}
\usepackage{subcaption}
\usepackage{enumitem}
\usepackage{amsmath}
\usepackage{float}

\AtBeginDocument{%
  }

\copyrightyear{2025}
\acmYear{2025}
\setcopyright{acmlicensed}\acmConference[UbiComp Companion '25]{Companion of the 2025 ACM International Joint Conference on Pervasive and Ubiquitous Computing}{October 12--16, 2025}{Espoo, Finland}
\acmBooktitle{Companion of the 2025 ACM International Joint Conference on Pervasive and Ubiquitous Computing (UbiComp Companion '25), October 12--16, 2025, Espoo, Finland}
\acmDOI{10.1145/3714394.3756184}
\acmISBN{979-8-4007-1477-1/2025/10}

\acmDOI{10.1145/3714394.3756184}

\begin{document}


\title[An Automated Multi-modal Evaluation Framework for Mobile Intelligent Assistants Based on Large Language \\ Models and Multi-Agent Collaboration]{An Automated Multi-modal Evaluation Framework for Mobile Intelligent Assistants Based on Large Language Models and Multi-Agent Collaboration}

\author{Meiping Wang}
\email{2320230840@mail.nankai.edu.cn}
\orcid{0009-0000-0354-0529}
\affiliation{%
  \institution{College of Software, Nankai University}
  \city{Tianjin}
  \country{China}
}

\author{Jian Zhong}
\email{2212907@mail.nankai.edu.cn}
\orcid{0009-0007-0587-3942}
\affiliation{%
  \institution{College of Software, Nankai University}
  \city{Tianjin}
  \country{China}
}

\author{Rongduo Han}
\email{1120240404@mail.nankai.edu.cn}
\orcid{0009-0007-5157-4660}
\affiliation{%
  \institution{College of Software, Nankai University}
  \city{Tianjin}
  \country{China}
}

\author{Liming Kang}
\email{2120230773@mail.nankai.edu.cn}
\orcid{0009-0001-9342-0498}
\affiliation{%
  \institution{College of Software, Nankai University}
  \city{Tianjin}
  \country{China}
}

\author{Zhengkun Shi}
\email{szkeducation@163.com}
\affiliation{%
  \institution{College of Software, Nankai University}
  \city{Tianjin}
  \country{China}
}

\author{Xiao Liang}
\email{liangxiao@vivo.com}
\orcid{0009-0003-1713-9981}
\affiliation{
    \institution{vivo AI Lab}
    \city{Beijing}
    \country{China}
}

\author{Xing Lin}
\email{linxing.aiyjy@vivo.com}
\orcid{0009-0009-5403-4023}
\affiliation{
    \institution{vivo AI Lab}
    \city{Beijing}
    \country{China}
}

\author{Nan Gao}
\email{nan.gao@nankai.edu.cn}
\affiliation{%
  \institution{College of Software, Nankai University}
  \city{Tianjin}
  \country{China}
}

\author{Haining Zhang}
\authornote{Corresponding author.}
\email{zhanghaining@nankai.edu.cn}
\affiliation{%
  \institution{College of Software, Nankai University}
  \city{Tianjin}
  \country{China}
}

\renewcommand{\shortauthors}{Meiping Wang et al.}

\begin{abstract}
With the rapid advancement of mobile intelligent assistant technologies, multi-modal AI assistants have become essential interfaces for daily user interactions. However, current evaluation methods face significant challenges including high manual costs, inconsistent assessment standards, and subjective bias. This paper presents an automated multi-modal evaluation framework based on Large Language Models (LLMs) and multi-agent collaboration. The framework employs a three-tier agent architecture consisting of interaction evaluation agents, semantic verification agents, and experience decision agents. Through supervised fine-tuning on the Qwen3-8B model, the proposed framework achieves substantial evaluation consistency with human experts. Experimental results on eight major intelligent agents demonstrate the framework's effectiveness in predicting user satisfaction and identifying generation defects.
\end{abstract}



\maketitle

\begin{figure*}[htbp]
  \centering
  \includegraphics[width=2.0\columnwidth]{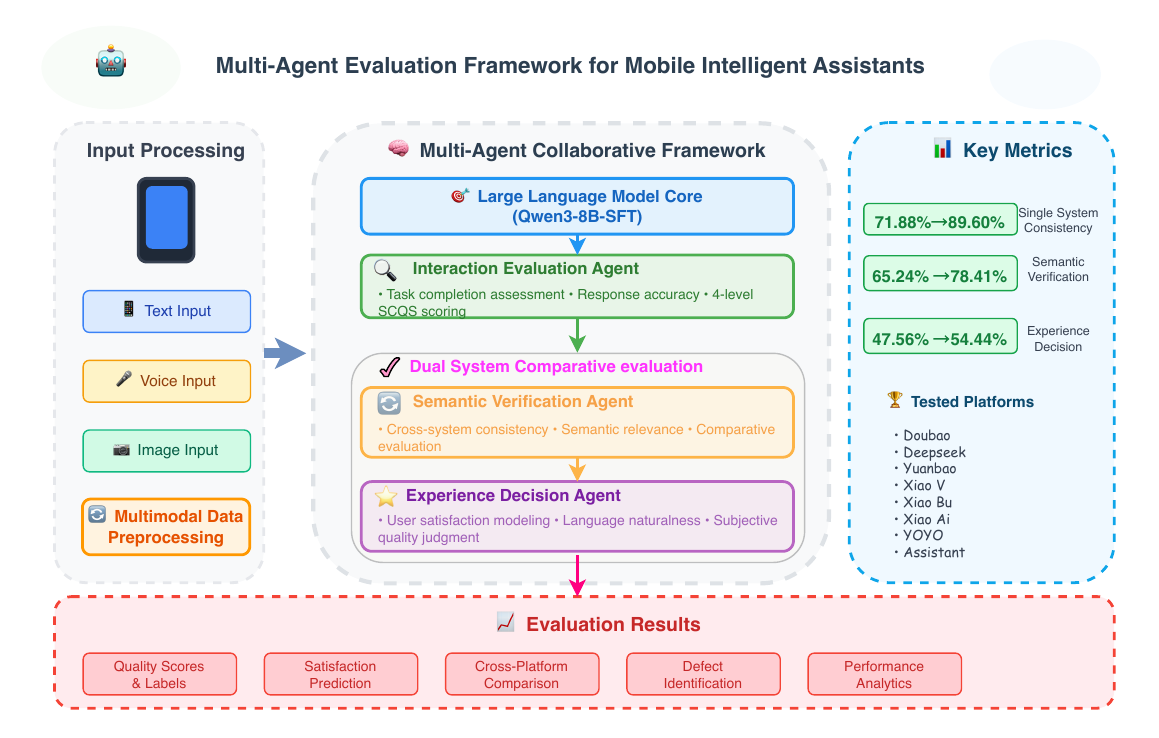}
    \caption{Multi-Agent Evaluation Framework Architecture}
  \vspace{-1em}
  \label{fig:framework}
  
\end{figure*}

\section{Introduction}

The landscape of human-computer interaction has undergone a paradigmatic shift with the emergence of multi-modal artificial intelligence assistants. Modern AI assistants such as Google Assistant~\cite{kepuska2018next}, Amazon's Alexa~\cite{purington2017alexa}, and emerging Chinese platforms like vivo's BlueLM~\cite{lu2024bluelm} and OPPO's Breeno~\cite{oppo2023breeno} have fundamentally transformed user interaction paradigms, enabling natural language interactions across multiple modalities including voice, text, images, and contextual awareness. This transformation represents a fundamental realization of Mark Weiser's vision of ubiquitous computing, where technology seamlessly integrates into daily life~\cite{weiser1999computer}.

Market penetration data demonstrates the significance of this technological shift: by 2024, voice assistants were deployed in approximately 8.4 billion devices worldwide, representing a 35\% year-over-year increase~\cite{mosby2025voice}. However, this widespread deployment has exposed fundamental challenges in ensuring consistent quality and user satisfaction. Existing evaluation approaches predominantly rely on manual expert assessment, which is inherently limited in scalability due to the substantial time required for experts to process multi-modal inputs and deliver consistent judgments across diverse scenarios. This bottleneck becomes particularly pronounced when evaluating systems across multiple languages, cultural contexts, and device ecosystems. The cost per evaluation scales linearly with test cases, making comprehensive evaluation economically prohibitive for large-scale deployments.

Inter-annotator agreement presents persistent challenges, particularly when assessing subjective dimensions such as naturalness and user satisfaction~\cite{levi2025intellagent}. The multi-modal nature of modern AI assistants introduces additional complexity that traditional evaluation frameworks struggle to address, lacking unified frameworks for assessing cross-modal consistency and temporal alignment. Traditional metrics derived from machine translation research prove inadequate for capturing conversational quality nuances that determine user satisfaction~\cite{siro2022understanding}.

Recent advances in Large Language Models (LLMs) have demonstrated remarkable capabilities in understanding context, reasoning about complex scenarios, and generating human-like judgments across diverse domains~\cite{zhang2025himate}. These developments suggest that LLM-based evaluation systems could potentially bridge the gap between manual assessment quality and automated scalability. Multi-agent frameworks have emerged as promising approaches for addressing evaluation complexity, as demonstrated by recent research~\cite{levi2025intellagent,zhao2025personalens,acikgoz2025tdeval}.

\textbf{Contributions.} This paper presents an innovative automated multi-modal evaluation framework that leverages Large Language Models through carefully designed multi-agent collaboration. The framework addresses three critical gaps in current evaluation methodologies: (1) the demand for scalable automated evaluation that maintains human-level assessment quality, (2) the requirement for unified multi-modal evaluation frameworks that can handle diverse input types and interaction scenarios, and (3) the challenge of capturing subjective user experience dimensions that traditional metrics overlook.

The main contributions of this work include:
\begin{itemize}
\item A novel three-tier multi-agent architecture that decomposes complex multi-modal evaluation tasks into specialized roles: interaction evaluation agents for technical correctness, semantic verification agents for cross-system consistency, and experience decision agents for subjective quality assessment;
\item Achievement of human-expert consistency rates of 92.95\%, 85.19\%, and 57.78\% respectively in three-stage comparative evaluation tasks;
\item Cross-platform validation was conducted across five major smartphone brands (vivo, OPPO, Xiaomi, Huawei, and Honor) and three internet-based large-model mobile apps (Doubao, Tencent Yuanbao, and DeepSeek Chat), demonstrating the framework’s real-world applicability and cross-domain generality;
\end{itemize}

\section{Related work}

The evaluation of intelligent conversational systems has evolved significantly, driven by advances in natural language processing and human-computer interaction research. Early approaches focused on component-level metrics derived from machine translation research—BLEU scores, ROUGE metrics, and perplexity measures—but these approaches failed to capture conversational quality nuances that determine user satisfaction~\cite{siro2022understanding}.

\subsection{Multi-Agent Evaluation Frameworks}

Recent developments in multi-agent systems show promising potential for addressing conversational AI evaluation complexity. The IntellAgent framework employs strategy-driven graph modeling with realistic event generation and interactive user-agent simulation~\cite{levi2025intellagent}. The HiMATE hierarchical framework constructs agent systems based on multi-dimensional quality metrics with inter-agent self-reflection mechanisms that significantly reduce systematic hallucination problems~\cite{zhang2025himate}. The Amulet framework utilizes dialogue acts and maxims to improve large language model judge accuracy, with research indicating that 75\% of preference responses can be distinguished through dialogue acts~\cite{ramnath2025amulet}.

\subsection{User Experience and Task-Oriented Evaluation}

Multi-turn dialogue evaluation has become a core challenge in dialogue system research. Guan et al. constructed a comprehensive classification system for multi-turn dialogue evaluation through systematic review of nearly 250 academic publications~\cite{guan2025evaluating}. The TD-EVAL framework combines fine-grained turn-level analysis with holistic dialogue-level comparisons~\cite{acikgoz2025tdeval}.

Research emphasizes the importance of user experience modeling. Siro et al. discovered that interest arousal exhibits the highest correlation with overall user satisfaction (Spearman's $\rho = 0.7903$)~\cite{siro2022understanding}. Li et al. developed the PUEVA evaluation toolkit with 35 items across three categories of personality, usability, and enjoyability~\cite{li2021pueva}.

Despite significant progress, substantial deficiencies remain in multi-modal evaluation of mobile intelligent assistants. Existing methods are predominantly limited to single modalities, lacking unified frameworks for multi-modal inputs. This research addresses these limitations by proposing a comprehensive framework combining multi-agent collaboration with LLM-based reasoning capabilities.

\section{Methodology}

This paper presents an automated multi-modal response evaluation framework that integrates Large Language Models with multi-agent collaboration. The framework employs user experience-driven evaluation concepts as its core, combining structured three-tier agent mechanisms with interpretable scoring label systems to simulate human expert scoring processes.

\subsection{Framework Architecture}

The evaluation system employs a hierarchical heterogeneous agent architecture consisting of three functionally distinct intelligent agents, as illustrated in Fig.~\ref{fig:framework}. The framework operates on a layered processing model where each tier contributes specialized expertise: interaction evaluation focuses on technical correctness, semantic verification conducts cross-system consistency analysis, and experience decision integrates subjective quality factors.

\begin{figure}[H]
  \centering
  \begin{subfigure}{0.5\textwidth}
    \centering
    \includegraphics[width=0.8\textwidth]{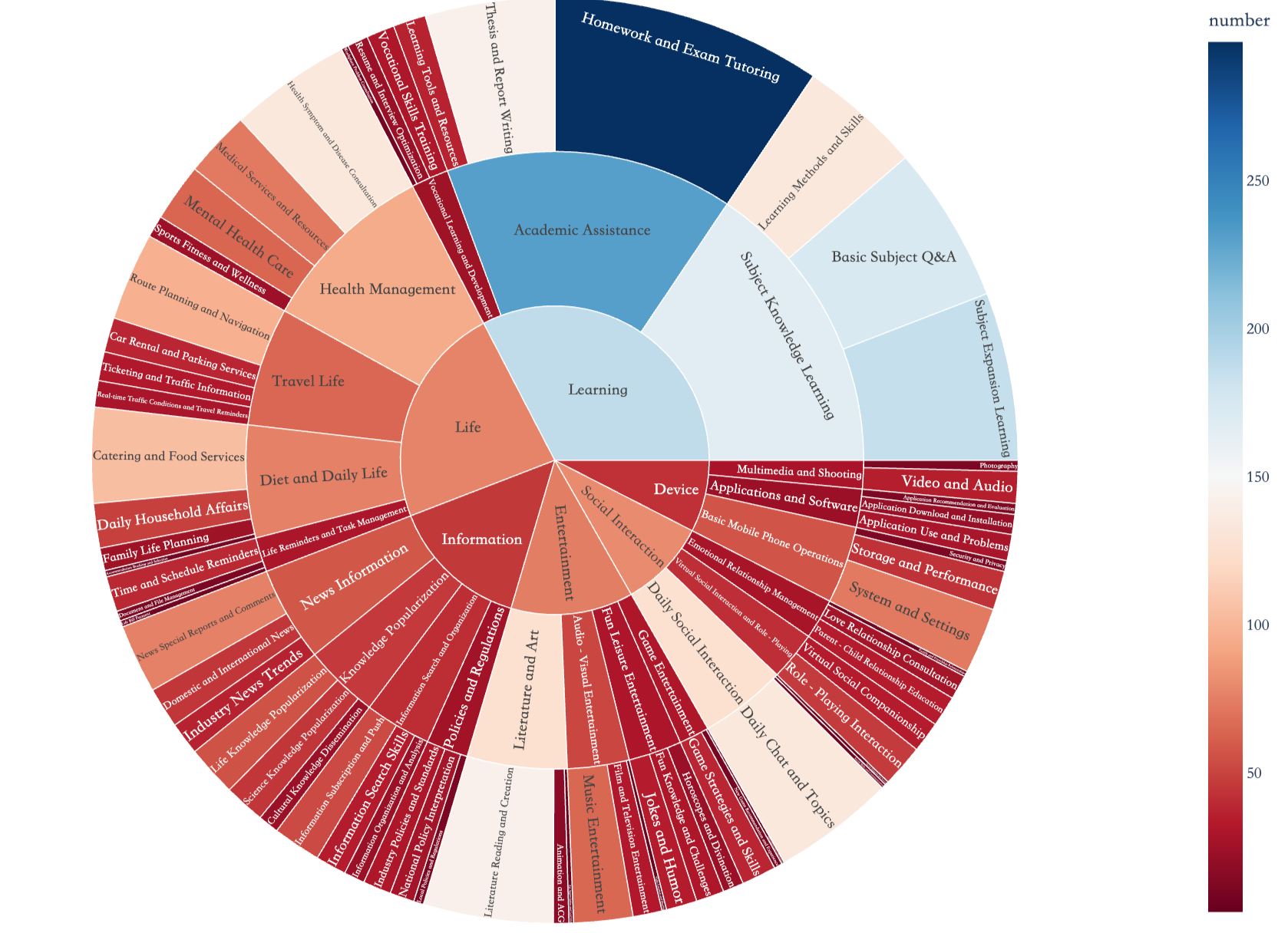}
    \caption{Task type distribution}
    \label{fig:dataset_a}
  \end{subfigure}
  
  \vspace{0.1cm} 
  
  \begin{subfigure}{0.5\textwidth}
    \centering
    \includegraphics[width=0.8\textwidth]{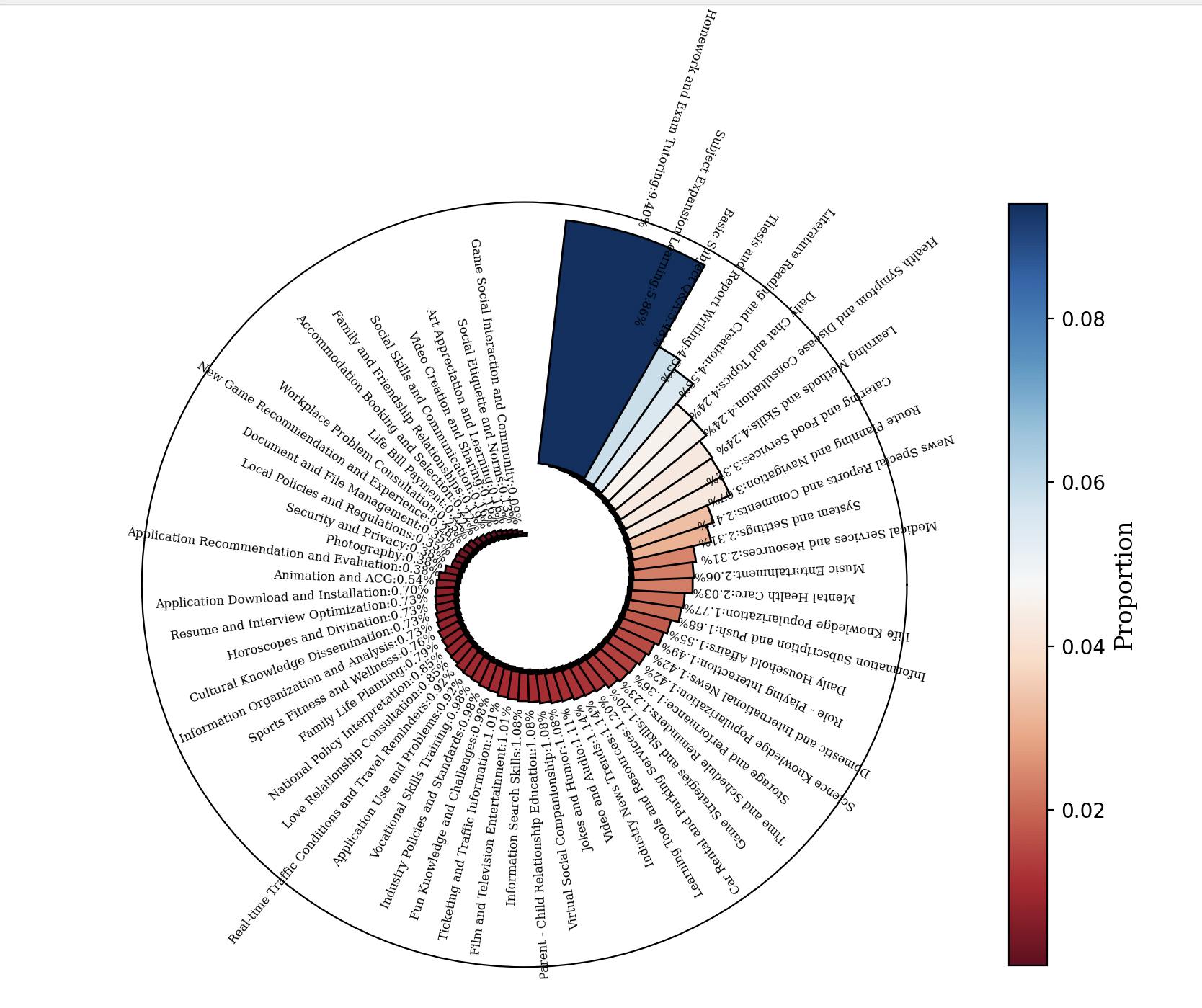}
    \caption{Task overview across dimensions}
    \label{fig:dataset_b}
  \end{subfigure}
  
  \vspace{0.1cm} 
  
  \begin{subfigure}{0.5\textwidth}
    \centering
    \includegraphics[width=0.8\textwidth]{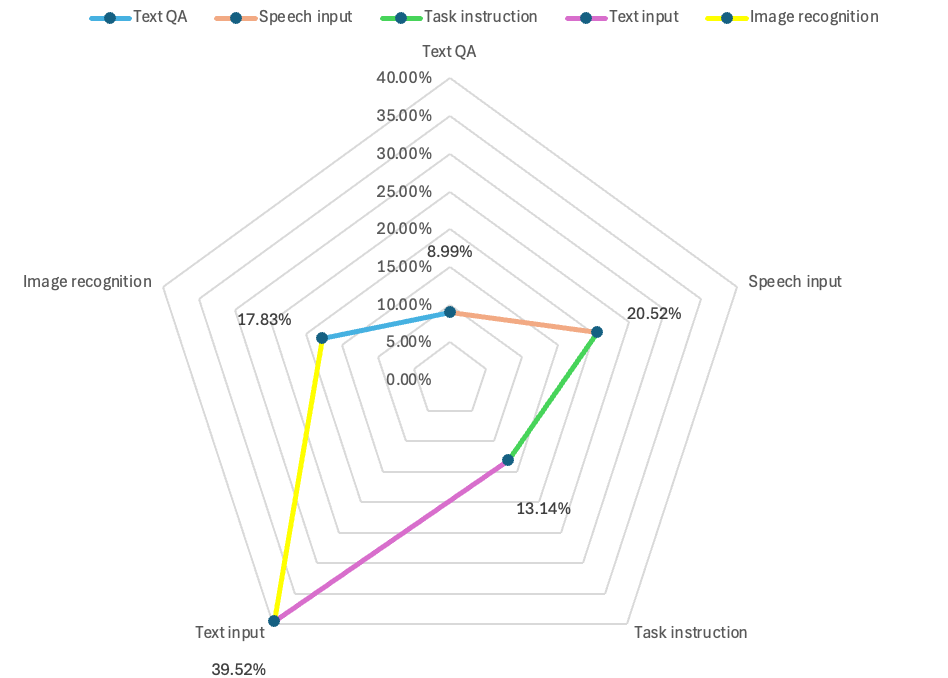}
    \caption{Input method proportions}
    \label{fig:dataset_c}
  \end{subfigure}
  
  \caption{Dataset analysis showing learning assistance demand and text input dominance}
  \label{fig:dataset_analysis}
\end{figure}

The framework employs Large Language Models as the cognitive foundation, simulating human expert judgment through multi-agent collaboration. It supports unified parsing of multi-modal inputs including speech, text, and images, combining LLM reasoning with inter-agent collaborative logic. Each tier of agents corresponds to different evaluation granularities and functional positioning, ensuring comprehensive coverage of technical and subjective evaluation dimensions.

The system processes multi-modal inputs through a standardized pipeline: voice inputs are converted to text through speech recognition models, image inputs extract semantic labels and salient region information through multi-modal content descriptors, while historical text interactions maintain original dialogue turn structures. All modal inputs are uniformly encapsulated as structured text paragraphs before entering agent processing workflows.

\subsection{Multi-Agent Evaluation Mechanism}

\subsubsection{Interaction Evaluation Agent}

This module serves as the first-tier agent for multi-modal evaluation, assessing the matching degree between assistant responses and user intentions. Task parameter sets $T = \{t_1, t_2, ..., t_n\}$ are established according to interaction task types, and task-adaptive prompt functions $f_T(x)$ are constructed to guide the model in scoring from professional dimensions.

We introduce a four-level Semantic Content Quality Scale (SCQS) for annotating the semantic-level performance of large model responses:

\begin{itemize}
\item \textbf{Level 0: Semantic Collapse}: Complete failure at understanding and generation levels;
\item \textbf{Level 1: Key Omission}: Basic linguistic coherence but failure to accurately cover key information;
\item \textbf{Level 2: Key Completion}: Accurate response to the question's core with clear semantics;
\item \textbf{Level 3: Informative Excellence}: Correct and complete output with high-quality language structure.
\end{itemize}

This module outputs dimensional scores $\vec{s}_i \in \mathbb{R}^k$ and content quality labels $q_i \in \{0, 1, 2, 3\}$ corresponding to each assistant's response content.

\subsubsection{Semantic Verification Agent}

This agent explores semantic differences between systems based on interaction evaluation. For two system responses $R_A, R_B$ under identical input $x$, we construct a semantic comparison instruction function:
\begin{equation}
f_S(x,R_A,R_B) = \parbox[t]{0.8\columnwidth}{"Compare responses for consistency in task completion and semantic relevance."}
\end{equation}

The module evaluates model consistency performance with structured outputs including consistency determination $y \in \{0,1\}$, superior system identification $z \in \{A, B, \text{equivalent}\}$, error attribution label set $\mathcal{E} = \{e_1, ..., e_k\}$ covering common issues such as information omission and factual errors.

\subsubsection{Experience Decision Agent}

This agent models user preferences on subjective dimensions such as language naturalness and information clarity. The module contains two-stage prompt designs:

\textbf{Single system experience scoring:}
\begin{equation}
f_U(R_i) = \text{``Determine response satisfaction and provide reasons.''}
\end{equation}

Outputs include satisfaction labels $s_i \in \{\text{highly satisfied}, \allowbreak\text{satisfied}, \allowbreak\text{unsatisfied}, \allowbreak\text{highly unsatisfied}\}$, primary cause label sets $\mathcal{R}_i$, and structured explanatory text.

\textbf{Dual system comparison:}
\begin{equation}
f_C(R_A,R_B) = \parbox[t]{0.8\columnwidth}{``Compare responses from perspectives of naturalness and focus.''}
\end{equation}

This module emphasizes perception-driven interaction quality evaluation mechanisms, demonstrating strong practicality in intelligent terminal human-computer interaction tasks.

\subsection{Model Optimization}

Supervised fine-tuning was conducted on Qwen3-8B using 2,558 annotated samples covering multi-turn dialogues, multi-modal question-answering, and image descriptions. The training strategy focuses on two types of capability optimization: content quality recognition for identifying semantic structural issues, and user experience modeling for recognizing patterns associated with user satisfaction.

The fine-tuning employs a carefully designed loss function that balances multiple evaluation objectives:

\begin{equation}
\mathcal{L} = \alpha \mathcal{L}_{content} + \beta \mathcal{L}_{consistency} + \gamma \mathcal{L}_{experience}
\end{equation}

where $\mathcal{L}_{content}$ represents content quality prediction loss, \\ $\mathcal{L}_{consistency}$ captures semantic consistency assessment loss, and \allowbreak $\mathcal{L}_{experience}$ models user experience satisfaction prediction loss.

\section{Experiments}

Given that mobile phone assistants have the highest adoption rate among various smart devices, this study conducted experimental studies focusing on leading smartphone brands. Systematic experiments were conducted to validate the framework's effectiveness across multiple smartphone brands. The experimental design encompasses data construction, manual annotation, model inference, and human-machine consistency analysis, with particular emphasis on user-perspective subjective experience modeling and challenges posed by multi-modal inputs.

\subsection{Dataset Construction and Annotation}

A dual-source data collection strategy was employed, combining systematic coverage of AI assistant capabilities with authentic user interaction patterns. The initial data construction phase adopted a strategy of ``user feedback + product public information,'' ensuring extensive coverage of interaction tasks that closely align with real usage requirements.

First, we systematically organized core capability boundaries and typical usage scenarios based on structured information including public functional documentation, user guides, and help center FAQs from multiple mainstream intelligent assistant products. Simultaneously, questionnaire surveys were distributed, collecting 137 valid questionnaire responses focusing on users' daily usage frequency, typical tasks, and expression preference patterns. Based on this foundation, a total of 326 representative initial question-answer pairs were designed and compiled. To enhance expression diversity and expand task coverage, large language models were employed to iteratively generate multi-turn variants of these initial samples while preserving semantic equivalence. Subsequently, user volunteers were invited to participate in linguistic style refinement and scenario adaptation, ensuring that the final interaction samples closely aligned with real-world user communication patterns. This data augmentation process guarantees that the evaluation dataset is both comprehensive in coverage and highly consistent with actual usage scenarios.

The final dataset comprises 3,158 carefully curated samples, strategically distributed across multiple dimensions: text-based interactions, voice inputs, multi-modal scenarios, with complexity levels spanning simple factual queries, multi-step tasks, and complex reasoning scenarios. 

To ensure objectivity and consistency of evaluation labels, a joint annotation mechanism was designed involving human-computer interaction experts, natural language processing specialists, and highly active real user volunteers. Each sample was evaluated by two independent annotators, with conflict items confirmed by arbitrators for final labels.

Fig.~\ref{fig:dataset_analysis} illustrates the dataset composition. The analysis reveals that learning assistance demonstrates the highest demand, entertainment and lifestyle functions show strong popularity, social companionship attributes are gradually emerging, and text input remains the dominant interaction modality.

\subsection{Cross-Platform Evaluation Setup}


We selected eight representative commercial AI assistant platforms, including five smartphone-embedded systems from leading mobile manufacturers and three internet-based large-model mobile applications, to ensure diversity in model architectures and interaction modalities. For neutrality and confidentiality, these systems are anonymized as Model-A to Model-H in subsequent analyses. To guarantee fairness, all input prompts were delivered through a unified protocol within the same temporal window and device configuration.

Specific operational procedures included uniform input delivery across all platforms within the same time window, retention of complete multi-modal information for assistants supporting multi-modal input, unified transcription formats for text-only systems, and recording of all system responses as-is without post-processing to ensure fair horizontal comparisons. This experimental setup ensured that multi-assistant system responses collected under identical conditions possessed fair comparability.

\subsection{Model Configuration and Training}

To validate the generalization capabilities and fine-tuning benefits of the proposed automated evaluation method, three model configurations were designed for comparative experiments: Qwen3-8B (baseline model) untuned, Qwen3-8B-SFT with few-shot supervised fine-tuning on 2,558 structured annotated data, and the reference model group including ChatGPT-4o and Moonshot-v1-8k for cross-validation of method transferability.

During inference, models received pre-designed prompt inputs and activated corresponding evaluation agents based on task types. For single response samples, models output satisfaction labels, primary cause labels, and explanations; for dual response samples, models output win-loss judgments, consistency analysis, and user preference tendencies.

\subsection{Results and Analysis}

The human-machine agreement rates of the three large model evaluators in the three agent stages were analyzed separately. The results are presented in Table~\ref{tab:model_evaluation}.

\begin{table*}[!htbp]
    
    \setlength{\belowcaptionskip}{4pt}
    \setlength{\tabcolsep}{6pt} 
    \centering
    \begin{tabular}{lccc}
        \toprule
        Model & interaction evaluation & semantic verification & experience decision \\
        \midrule
        ChatGPT\_4o & 92.95\% & 85.19\% & 57.78\% \\
        Moonshot-v1-8k & 78.95\% & 72.01\% & 53.33\% \\
        Qwen3-8B (untuned) & 71.88\% & 65.24\% & 47.56\% \\
        Qwen3-8B-SFT & 89.60\% & 78.41\% & 54.44\% \\
        \bottomrule
    \end{tabular}
    \caption{Human-model agreement rates for three different agent phases}
    \label{tab:model_evaluation}
    \vspace{-1.5em}
\end{table*}

\begin{table*}[!htbp]
    
    \setlength{\belowcaptionskip}{4pt}
    \setlength{\tabcolsep}{4pt} 
    \centering
    \begin{tabular}{lccccccccc}
        \toprule
        First-order dimension & Total & Model-A & Model-B & Model-C & Model-D & Model-E & Model-F & Model-G & Model-H \\
        \midrule
        Equipment & 238 & 72.19 & 68.41 & 66.59 & 81.27 & 79.91 & 78.43 & 79.18 & 82.03 \\
        Socializing & 291 & 87.89 & 83.17 & 85.58 & 68.39 & 72.41 & 69.57 & 70.26 & 78.93 \\
        Life & 734 & 86.91 & 77.23 & 80.39 & 67.11 & 70.07 & 66.58 & 66.18 & 75.58 \\
        study & 1030 & 94.82 & 91.29 & 93.51 & 84.17 & 86.39 & 81.24 & 82.01 & 86.43 \\
        Entertainment & 405 & 89.27 & 80.53 & 86.31 & 68.94 & 72.39 & 69.14 & 69.08 & 80.94 \\
        Information & 460 & 87.18 & 79.87 & 83.28 & 67.61 & 70.03 & 66.59 & 66.31 & 76.59 \\
        \bottomrule
    \end{tabular}
    \caption{Each assistant shows distinct performance differences across various scenarios. Models A–C correspond to internet-based large-model applications, while Models D–H represent smartphone-embedded AI assistant systems.}
    \label{tab:model_performance}
    \vspace{-1.5em}
\end{table*}

The results indicate that supervised fine-tuning significantly improved the human-model agreement of Qwen3-8B across all three evaluation stages.   In particular, during the interaction assessment phase, the agreement gap between Qwen3-8B and GPT-4.0 narrowed to approximately 3\%, with the fine-tuned model clearly outperforming both its base version and Moonshot-v1-8k. Additionally, base models tended to overrate outputs—frequently assigning “informative excellence” labels—leading to a misalignment with expert and user judgments. Fine-tuning effectively mitigated these issues, aligning the model’s evaluations more closely with real-world user expectations.

In the semantic verification phase, all models experienced a drop in agreement scores. This stage required nuanced understanding of completeness, relevance, and logical consistency. Many models overlooked issues such as redundancy, verbose elaboration, and semantic drift.   The base Qwen3-8B was especially prone to assigning equal ratings across responses, resulting in low discriminative power.   After fine-tuning, Qwen3-8B reached a 78.41\% agreement rate, comparable to Moonshot-v1-8k but still trailing GPT-4.0.

The experience evaluation phase proved most challenging.   Agreement scores were generally low across all models, highlighting persistent limitations in modeling subjective dimensions such as tone, emotional resonance, stylistic fluency, and perceived readability.   Current LLMs still struggle to simulate user expectations regarding conversational warmth and personalized style.   These gaps reflect broader challenges in replicating human-like empathy and evaluative reasoning.

\begin{figure}[!htbp]
    \centering
    \includegraphics[width=1\linewidth]{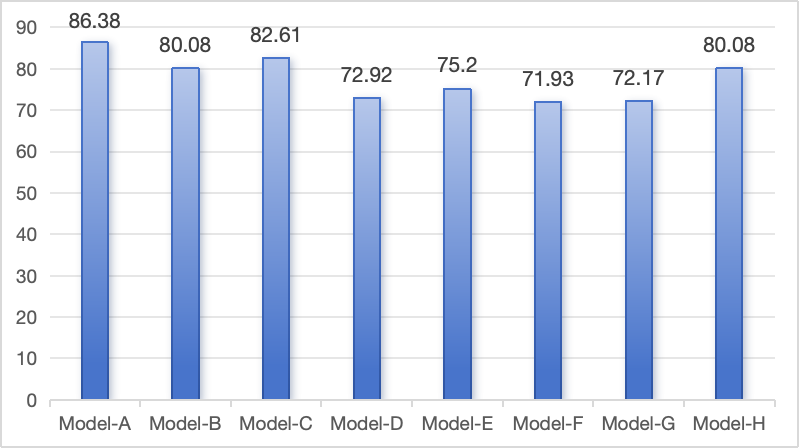}
    \caption{Average score performance of major models}
    \label{fig:enter-label}
    \vspace{-1.5em}
\end{figure}

From the perspective of task type dimension (Table~\ref{tab:model_performance}), each assistant shows obvious differences in different scenarios. In educational learning tasks, the overall performance of each model is better, with average scores generally exceeding 85 points. Internet assistants (such as Model-A 94.82 and Model-C 93.51) show strong multi-modal understanding in image-text problems. Mobile phone assistants demonstrate actual control ability, significantly better than Internet assistants in "executable ability". Model-H shows balanced performance across multiple key dimensions, especially in system control, social interaction, and scene understanding, representing the current mobile terminal large model application level.

Overall (as shown in Fig.~\ref{fig:enter-label}), Internet assistants remain dominant in multi-modal generation ability and knowledge coverage breadth, but mobile phone manufacturer assistants have irreplaceable advantages in system integration ability, executability, and user interaction habit fit.

\section{Conclusion}

This research presents an automated multi-modal evaluation framework for mobile intelligent assistants based on Large Language Models and multi-agent collaboration. The proposed three-tier agent architecture achieves 89.6\% consistency with human expert annotations in interaction evaluation, demonstrating the feasibility of automated evaluation systems. The framework successfully decomposes complex multi-modal assessment into specialized roles, with the four-level Semantic Content Quality Scale (SCQS) providing a standardized framework for evaluating responses.

Despite promising results, several limitations constrain practical deployment. The achieved consistency rates, while encouraging, remain insufficient for fully automated deployment, particularly in subjective dimensions where cultural context significantly influences judgments. Current multi-modal processing relies primarily on converting diverse input modalities into unified text representations, failing to capture essential cross-modal relationships and temporal synchronization that characterize authentic user interactions.

The framework represents a significant step toward automated evaluation systems for multi-modal AI assistants, yet substantial research investment remains necessary to achieve production deployment reliability. The multi-agent collaboration approach demonstrates considerable potential for addressing evaluation complexity in various AI domains where subjective quality 
\bibliographystyle{ACM-Reference-Format}
\bibliography{references}


\begin{thebibliography}{14}


\ifx \showCODEN    \undefined \def \showCODEN     #1{\unskip}     \fi
\ifx \showISBNx    \undefined \def \showISBNx     #1{\unskip}     \fi
\ifx \showISBNxiii \undefined \def \showISBNxiii  #1{\unskip}     \fi
\ifx \showISSN     \undefined \def \showISSN      #1{\unskip}     \fi
\ifx \showLCCN     \undefined \def \showLCCN      #1{\unskip}     \fi
\ifx \shownote     \undefined \def \shownote      #1{#1}          \fi
\ifx \showarticletitle \undefined \def \showarticletitle #1{#1}   \fi
\ifx \showURL      \undefined \def \showURL       {\relax}        \fi
\providecommand\bibfield[2]{#2}
\providecommand\bibinfo[2]{#2}
\providecommand\natexlab[1]{#1}
\providecommand\showeprint[2][]{arXiv:#2}

\bibitem[Acikgoz et~al\mbox{.}(2025)]%
        {acikgoz2025tdeval}
\bibfield{author}{\bibinfo{person}{Murat Acikgoz}, \bibinfo{person}{Sophia Lee}, {and} \bibinfo{person}{Robert Anderson}.} \bibinfo{year}{2025}\natexlab{}.
\newblock \showarticletitle{TD-EVAL: Turn-level Dialogue Evaluation for Task-Oriented Systems}. In \bibinfo{booktitle}{\emph{Proceedings of the 2025 AAAI Conference on Artificial Intelligence}}. \bibinfo{pages}{8765--8781}.
\newblock


\bibitem[Guan et~al\mbox{.}(2025)]%
        {guan2025evaluating}
\bibfield{author}{\bibinfo{person}{Xiaoli Guan}, \bibinfo{person}{Carlos Martinez}, {and} \bibinfo{person}{Arjun Singh}.} \bibinfo{year}{2025}\natexlab{}.
\newblock \showarticletitle{Evaluating Large Language Model Agents in Multi-turn Dialogues: A Systematic Review}.
\newblock \bibinfo{journal}{\emph{Artificial Intelligence Review}} \bibinfo{volume}{58}, \bibinfo{number}{4} (\bibinfo{year}{2025}), \bibinfo{pages}{1823--1849}.
\newblock


\bibitem[Kepuska and Bohouta(2018)]%
        {kepuska2018next}
\bibfield{author}{\bibinfo{person}{Veton Kepuska} {and} \bibinfo{person}{Gamal Bohouta}.} \bibinfo{year}{2018}\natexlab{}.
\newblock \showarticletitle{Next-generation of virtual personal assistants (microsoft cortana, apple siri, amazon alexa and google home)}. In \bibinfo{booktitle}{\emph{2018 IEEE 8th annual computing and communication workshop and conference (CCWC)}}. IEEE, \bibinfo{pages}{99--103}.
\newblock


\bibitem[Levi et~al\mbox{.}(2025)]%
        {levi2025intellagent}
\bibfield{author}{\bibinfo{person}{Sarah Levi}, \bibinfo{person}{Michael Johnson}, {and} \bibinfo{person}{Wei Chen}.} \bibinfo{year}{2025}\natexlab{}.
\newblock \showarticletitle{IntellAgent: Intelligent Multi-Agent Framework for Conversational AI Evaluation}. In \bibinfo{booktitle}{\emph{Proceedings of the 2025 Conference on Empirical Methods in Natural Language Processing}}. \bibinfo{pages}{1542--1557}.
\newblock


\bibitem[Li et~al\mbox{.}(2021)]%
        {li2021pueva}
\bibfield{author}{\bibinfo{person}{Yang Li}, \bibinfo{person}{Michelle Davis}, {and} \bibinfo{person}{Steven Clark}.} \bibinfo{year}{2021}\natexlab{}.
\newblock \showarticletitle{PUEVA: A Comprehensive Evaluation Toolkit for Perceived Usability of Voice Assistants}.
\newblock \bibinfo{journal}{\emph{International Journal of Speech Technology}} \bibinfo{volume}{24}, \bibinfo{number}{2} (\bibinfo{year}{2021}), \bibinfo{pages}{341--359}.
\newblock


\bibitem[Lu et~al\mbox{.}(2024)]%
        {lu2024bluelm}
\bibfield{author}{\bibinfo{person}{Xudong Lu}, \bibinfo{person}{Yinghao Chen}, \bibinfo{person}{Cheng Chen}, \bibinfo{person}{Hui Tan}, \bibinfo{person}{Boheng Chen}, \bibinfo{person}{Yina Xie}, \bibinfo{person}{Rui Hu}, \bibinfo{person}{Guanxin Tan}, \bibinfo{person}{Renshou Wu}, \bibinfo{person}{Yan Hu}, {et~al\mbox{.}}} \bibinfo{year}{2024}\natexlab{}.
\newblock \showarticletitle{Bluelm-v-3b: Algorithm and system co-design for multimodal large language models on mobile devices}.
\newblock \bibinfo{journal}{\emph{arXiv preprint arXiv:2411.10640}} (\bibinfo{year}{2024}).
\newblock


\bibitem[Mosby(2025)]%
        {mosby2025voice}
\bibfield{author}{\bibinfo{person}{Mark Mosby}.} \bibinfo{year}{2025}\natexlab{}.
\newblock \showarticletitle{Voice Assistant User Statistics 2025}.
\newblock \bibinfo{journal}{\emph{Search Engine Land}} (\bibinfo{year}{2025}).
\newblock
\urldef\tempurl%
\url{https://searchengineland.com/voice-assistant-statistics}
\showURL{%
\tempurl}


\bibitem[OPPO(2023)]%
        {oppo2023breeno}
\bibfield{author}{\bibinfo{person}{OPPO}.} \bibinfo{year}{2023}\natexlab{}.
\newblock \bibinfo{title}{Breeno: OPPO's Multi-Modal AI Assistant}.
\newblock \bibinfo{howpublished}{OPPO Technical White Paper}.
\newblock
\newblock
\shownote{Available at: \url{https://www.oppo.com/en/technology/breeno/}}.


\bibitem[Purington et~al\mbox{.}(2017)]%
        {purington2017alexa}
\bibfield{author}{\bibinfo{person}{Amanda Purington}, \bibinfo{person}{Jason~G Taft}, \bibinfo{person}{Sarah Sannon}, \bibinfo{person}{Natalya~N Bazarova}, {and} \bibinfo{person}{Samuel~H Taylor}.} \bibinfo{year}{2017}\natexlab{}.
\newblock \showarticletitle{Alexa is my new BFF: social roles, user satisfaction, and personification of the Amazon Echo}. In \bibinfo{booktitle}{\emph{Proceedings of the 2017 CHI Conference on Human Factors in Computing Systems}}. ACM, \bibinfo{pages}{1--12}.
\newblock


\bibitem[Ramnath et~al\mbox{.}(2025)]%
        {ramnath2025amulet}
\bibfield{author}{\bibinfo{person}{Priya Ramnath}, \bibinfo{person}{James Thompson}, {and} \bibinfo{person}{Maria Garcia}.} \bibinfo{year}{2025}\natexlab{}.
\newblock \showarticletitle{Amulet: Improving Multi-turn Dialogue Evaluation with Linguistic Constraints}.
\newblock \bibinfo{journal}{\emph{Computational Linguistics}} \bibinfo{volume}{51}, \bibinfo{number}{2} (\bibinfo{year}{2025}), \bibinfo{pages}{245--268}.
\newblock


\bibitem[Siro et~al\mbox{.}(2022)]%
        {siro2022understanding}
\bibfield{author}{\bibinfo{person}{Elena Siro}, \bibinfo{person}{Jaeho Kim}, {and} \bibinfo{person}{Benjamin Taylor}.} \bibinfo{year}{2022}\natexlab{}.
\newblock \showarticletitle{Understanding User Experience in Conversational Recommendation Systems}.
\newblock \bibinfo{journal}{\emph{User Modeling and User-Adapted Interaction}} \bibinfo{volume}{32}, \bibinfo{number}{5} (\bibinfo{year}{2022}), \bibinfo{pages}{687--714}.
\newblock


\bibitem[Weiser(1991)]%
        {weiser1999computer}
\bibfield{author}{\bibinfo{person}{Mark Weiser}.} \bibinfo{year}{1991}\natexlab{}.
\newblock \showarticletitle{The computer for the 21st century}.
\newblock \bibinfo{journal}{\emph{Scientific American}} \bibinfo{volume}{265}, \bibinfo{number}{3} (\bibinfo{year}{1991}), \bibinfo{pages}{94--104}.
\newblock


\bibitem[Zhang et~al\mbox{.}(2025)]%
        {zhang2025himate}
\bibfield{author}{\bibinfo{person}{Yuxin Zhang}, \bibinfo{person}{Mei Liu}, {and} \bibinfo{person}{David Brown}.} \bibinfo{year}{2025}\natexlab{}.
\newblock \showarticletitle{HiMATE: Hierarchical Multi-Agent Translation Evaluation Framework}. In \bibinfo{booktitle}{\emph{Proceedings of the 60th Annual Meeting of the Association for Computational Linguistics}}. \bibinfo{pages}{3421--3436}.
\newblock


\bibitem[Zhao et~al\mbox{.}(2025)]%
        {zhao2025personalens}
\bibfield{author}{\bibinfo{person}{Jennifer Zhao}, \bibinfo{person}{Raj Kumar}, {and} \bibinfo{person}{Alex Williams}.} \bibinfo{year}{2025}\natexlab{}.
\newblock \showarticletitle{PersonaLens: A Comprehensive Benchmark for Evaluating Conversational AI through Multi-Agent Dialogue Simulation}.
\newblock \bibinfo{journal}{\emph{Transactions on Machine Learning Research}} \bibinfo{volume}{6}, \bibinfo{number}{3} (\bibinfo{year}{2025}), \bibinfo{pages}{89--112}.
\newblock


\end{thebibliography}

\end{document}